\documentclass{article}

\usepackage[final]{nips_2018}

\usepackage[utf8]{inputenc} 
\usepackage[T1]{fontenc}    
\usepackage{hyperref}       
\usepackage{url}            
\usepackage{booktabs}       
\usepackage{amsfonts}       
\usepackage{nicefrac}       
\usepackage{microtype}      
\usepackage{graphicx,url,color}
\usepackage{amsmath,amssymb,amsopn,amsthm,bm,bbm}
\usepackage[textsize=small]{todonotes}
\usepackage{placeins}
\usepackage{multicol}
\usepackage{multirow,rotating}

\newcommand{\D}{\mathcal{D}}

\title{Counterfactual Learning from Human Proofreading Feedback for Semantic Parsing}
\author{Carolin Lawrence \\
	Computational Linguistics,\\Heidelberg University,\\Germany \\\texttt{lawrence@cl.uni-heidelberg.de}\\\And
	Stefan Riezler\\
	Computational Linguistics \& IWR,\\Heidelberg University,\\Germany\\\texttt{riezler@cl.uni-heidelberg.de}}

\begin{document}


\maketitle

\begin{abstract}
In semantic parsing for question-answering, it is often too expensive to collect gold parses or even gold answers as supervision signals. We propose to convert model outputs into a set of human-understandable statements which allow non-expert users to act as proofreaders, providing error markings as learning signals to the parser. Because model outputs were suggested by a historic system, we operate in a counterfactual, or off-policy, learning setup. We introduce new estimators which can effectively leverage the given feedback and which avoid known degeneracies in counterfactual learning, while still being applicable to stochastic gradient optimization for neural semantic parsing. Furthermore, we discuss how our feedback collection method can be seamlessly integrated into deployed virtual personal assistants that embed a semantic parser. Our work is the first to show that semantic parsers can be improved significantly by counterfactual learning from logged human feedback data.\footnote{Parts of the work in this paper have been previously published in the Proceedings of the 56th Annual Meeting of the Association for Computational Linguistics (ACL) \citep{LawrenceRiezler:18}.}

\end{abstract}

\section{Introduction} 

\begin{figure*}[h]
	\centering
	\includegraphics[width=1.0\textwidth,keepaspectratio]{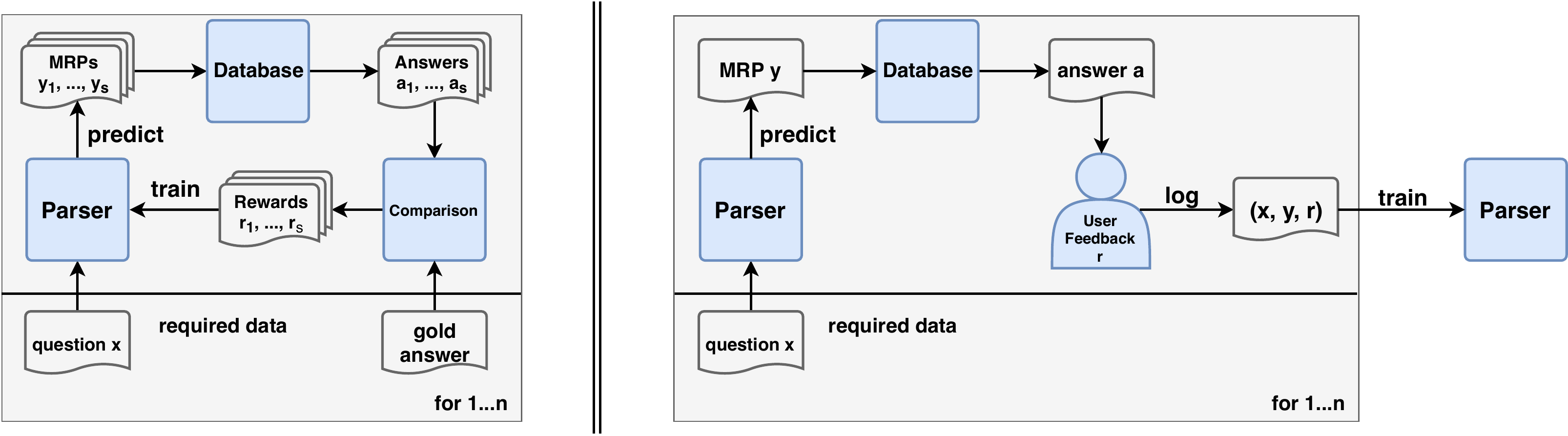}
	\caption{Left: Online reinforcement learning setup for semantic parsing  where both questions and gold answers are available. The parser attempts to find correct machine readable parses (MRPs) by producing multiple parses, obtaining corresponding answers,  and comparing them against the gold answer.
		Right: In our setup, a question does not have an associated gold answer. The parser outputs a single MRP and the corresponding answer is shown to a user who provides some feedback. Such triplets are collected in a log which can be used for offline training of a semantic parser. This scenario is called counterfactual since the feedback was logged for outputs from a system different from the target system to be optimized.}
	\label{fig:diagram}
\end{figure*}

Recent work (\citet{LiangETAL:17,MouETAL:17a,PengETAL:17}; \textit{inter alia}) has applied reinforcement learning to address the annotation bottleneck in semantic parsing as follows: Given a question, the existence of a corresponding gold answer is assumed. A semantic parser produces multiple parses per question and corresponding answers are obtained. These answers are then compared against the gold answer and a positive reward is recorded if there is an overlap. The parser is then guided towards correct parses using the REINFORCE algorithm \citep{Williams:92} which scales the gradient for the various parses by their obtained reward (see the left half of Figure \ref{fig:diagram}). However, learning from question-answer pairs is only efficient if gold answers exist and are cheap to obtain. For complex open-domain question-answering tasks, correct answers are not unique factoids, but open-ended lists, counts in large ranges, or fuzzily defined objects. For example, geographical queries against databases such as OpenStreetMap (OSM) can involve fuzzy operators such as ``near'' or ``in walking distance'' and thus need to allow for fuzziness in the answers as well.

A possible solution lies in machine learning from even weaker supervision signals in form of human bandit feedback\footnote{The term ``bandit feedback'' is inspired by the scenario of maximizing the reward for a sequence of pulls of arms of ``one-armed bandit'' slot machines.} where the semantic parsing system suggests exactly one parse for which feedback is collected from a human user. In this setup neither gold parse nor gold answer are known and feedback is obtained for only one system output per question. In this paper, we propose to convert system outputs into sets of human-understandable statements which allow non-expert users to act as proofreaders who mark errors in the predicted parse. These error markings can be used in a counterfactual, or off-policy, learning scenario to improve the parsing model \citep{BottouETAL:13} (see the right half of Figure \ref{fig:diagram}).

Our user interface allows to collect feedback based on the parse rather than the answer by automatically converting a parse to a set of statements that can be marked as correct or incorrect (see Figure \ref{fig:user_interface2}). From a reinforcement learning perspective \citep{SuttonBarto:98}, this approach corresponds to factorizing rewards at the token level, but performing off-policy learning instead of online updates as in actor-critic methods \citep{KondaTsitsiklis:00}.  In difference to imitation learning \citep{RossETAL:11}, users do not have to provide correct actions, but just have to mark erroneous tokens. We show that users can provide such token-level feedback for one whole parse in 16.4 seconds on average. This exemplifies that our approach is more efficient and cheaper than recruiting experts to annotate parses or asking workers to compile large answer sets.

Our experiments show that counterfactual learning can be applied to neural sequence-to-sequence learning for semantic parsing. A baseline neural semantic parser is trained in fully supervised fashion, human bandit feedback from human users is collected in a log and subsequently used to improve the parser. The resulting parser significantly outperforms the baseline model as well as a simple bandit-to-supervised approach (B2S) where the subset of completely correct parses is treated as a supervised dataset. Even with feedback to only 995 model outputs, the baseline system is improved by about 1 percentage point in answer F1 score without ever seeing a gold standard parse.
Finally, we repeat our experiments on a larger but simulated log to show that our gains can scale: the baseline system is now improved by 7.45 points. This presents a promising result for commercial virtual personal assistants that can easily collect large amounts of feedback.
\section{Neural Semantic Parsing}\label{sec:semparse}

Our semantic parsing model is a state-of-the-art sequence-to-sequence neural network using an encoder-decoder setup \citep{ChoETAL:14,SutskeverETAL:14} together with an attention mechanism \citep{BahdanauETAL:15}. 
The network maps an input sequence $x = x_1, x_2, \dots x_{|x|}$ to an output sequence $y = y_1, y_2, \dots y_{|y|}$ with probability
%

\begin{equation}
\pi_w(y|x) = \prod_{j=1}^{|y|} \pi_w(y_j|y_{<j},x),
\end{equation}
where $y_j$ are the individual tokens of $y$.

In our case, output sequences are linearized machine readable parses, called queries in the following. Given supervised data $\D_{sup} = \{(x_t,\bar{y}_t)\}_{t=1}^n$ of question-query pairs, where $\bar{y}_t$ is the true target query for $x_t$, the neural network can be trained using SGD and a cross-entropy (CE) objective. With $y_{t,<j} = y_{t,1}, y_{t,2} \dots y_{t,j-1}$, we define:

\begin{equation}
\mathcal{L}_{CE} = - \frac{1}{n} \sum_{t=1}^{n} \sum_{j=1}^{|\bar{y}|} \log \pi_w(\bar{y}_{t,j} | \bar{y}_{t,<j}, x_t).
\end{equation}

\section{Counterfactual Learning for Semantic Parsing}

\paragraph{Counterfactual Learning Objectives.} 
We assume a policy $\pi_w$ that, given an input $x \in \mathcal{X}$, defines a conditional probability distribution over possible outputs $y \in \mathcal{Y}(x)$. Furthermore, we assume that the policy is parameterized by $w$ and its gradient can be derived.  We also assume that the model decomposes over individual output tokens, i.e. that the model produces the output token by token.

The counterfactual learning problem can be described as follows: We are given a data log of triples $\D_{log}=\{(x_t,y_t,\delta_t)\}_{t=1}^n$ where outputs $y_t$ for inputs $x_t$ were generated by a logging system under policy $\pi_0$, and loss values $\delta_t \in [-1,0]$\footnote{We use the terms loss and (negative) rewards interchangeably, depending on context.} were observed for the generated data points. Our goal is to optimize the expected reward (in our case: minimize the expected risk) for a target policy $\pi_w$ given the data log $\D_{log}$.
In case of deterministic logging, outputs are logged with propensity $\pi_0(y_t|x_t) = 1, \; t=1, \dots, n$. This results in a \emph{deterministic propensity matching (DPM)} objective \citep{LawrenceETAL:17}, without the possibility to correct the sampling bias of the logging policy by inverse propensity scoring \citep{RosenbaumRubin:83}:
\begin{align}
\label{eq:emp-risk} 
\hat{R}_{\text{DPM}}(\pi_w) = \frac{1}{n} \sum_{t=1}^n \delta_t \pi_w(y_t|x_t).
\end{align}

This objective can show degenerate behavior in that it overfits to the choices of the logging policy \citep{SwaminathanJoachimsNIPS:15,LawrenceETALnips:17}. This degenerate behavior can be avoided by reweighting using a multiplicative control variate \citep{Kong:92,PrecupETAL:00,JiangLi:16,ThomasBrunskill:16}. The new objective is called the \emph{reweighted deterministic propensity matching (DPM+R)} objective in \citet{LawrenceETAL:17}:
\begin{align}
\label{eq:r} 
\hat{R}_{\text{DPM+R}}(\pi_w) &= \frac{1}{n} \sum_{t=1}^{n} \delta_t \bar{\pi}_w(y_t|x_t) \\ \notag
&= \frac{\frac{1}{n}\sum_{t=1}^{n} \delta_t \pi_w(y_t|x_t)}{\frac{1}{n}\sum_{t=1}^{n} \pi_w(y_t|x_t)}.
\end{align}

\paragraph{Reweighting in Stochastic Learning.} As shown in \citet{SwaminathanJoachimsNIPS:15} and \citet{LawrenceETALnips:17}, reweighting over the entire data log $\D_{log}$ is crucial since it avoids that high loss outputs in the log take away probability mass from low loss outputs. This multiplicative control variate has the additional effect of reducing the variance of the estimator, at the cost of introducing a bias of order $O(\frac{1}{n})$ that decreases as $n$ increases \citep{Kong:92}. The desirable properties of this control variate cannot be realized in a stochastic (minibatch) learning setup since minibatch sizes large enough to retain the desirable reweighting properties are infeasible for large neural networks.

We offer a simple solution to this problem that nonetheless retains all desired properties of the reweighting. The idea is inspired by one-step-late algorithms that have been introduced for EM algorithms \citep{Green:90}. In the EM case, dependencies in objectives are decoupled by evaluating certain terms under parameter settings from previous iterations (thus: one-step-late) in order to achieve closed-form solutions. In our case, we decouple the reweighting from the parameterization of the objective by evaluating the reweighting under parameters $w'$ from some previous iteration. This allows us to perform gradient descent updates and reweighting asynchronously. Updates are performed using minibatches, however, reweighting is based on the entire log, allowing us to retain the desirable properties of the control variate.

The new objective, called \textit{one-step-late reweighted DPM} objective (DPM+OSL), optimizes $\pi_{w,w'}$ with respect to $w$ for a minibatch of size $m$, with reweighting over the entire log of size $n$ under parameters $w'$:
\begin{align}
\label{eq:prob_mini} 
\hat{R}_{\text{DPM+OSL}}(\pi_w) &= \frac{1}{m}\sum_{t=1}^{m} \delta_t \bar{\pi}_{w,w'}(y_t|x_t) \\ \notag
&= \frac{\frac{1}{m}\sum_{t=1}^{m} \delta_t \pi_w(y_t|x_t)}{\frac{1}{n}\sum_{t=1}^{n} \pi_{w'}(y_t|x_t)}.
\end{align}
If the renormalization is updated periodically, e.g. after every validation step, renormalizations under $w$ or $w'$ are not much different and will not hamper convergence. Despite losing the formal justification from the perspective of control variates, we found empirically that the OSL update schedule for reweighting is sufficient and does not deteriorate performance.

\paragraph{Token-Level Rewards.} For our application of counterfactual learning to human bandit feedback, we found another deviation from standard counterfactual learning to be helpful: For humans, it is hard to assign a graded reward to a query at a sequence level because either the query is correct or it is not. Furthermore, non-expert users cannot judge a parse because they are unfamiliar with the underlying machine readable language. Also, with a sequence level reward of 0 for incorrect queries, we do not know which part of the query is wrong and which parts might be correct. Assigning rewards at token-level will ease the feedback task and allow the semantic parser to learn from partially correct queries. 
For this, we move to $\log$ probabilities, which allows us to decompose the sequence into a sum over individual tokens.

Now tokens with positive feedback can be encouraged while tokens with negative feedback are ignored.
Thus, assuming the underlying policy can decompose over tokens, a token level (DPM+T) reward objective can be defined as follows:

\begin{align}
\label{eq:token} 
\hat{R}_{\text{DPM+T}}(\pi_w) &=\frac{1}{n} \sum_{t=1}^{n} \left(\sum_{j=1}^{|y|} \delta_{t,j} \log \pi_w(y_{t,j}|y_{t,<j}, x_t)\right).
\end{align}

Analogously, we can define an objective that combines the token-level rewards and the minibatched reweighting (DPM+T+OSL):
\begin{align}
\label{eq:token_prob} 
\hat{R}_{\text{DPM+T+OSL}}(\pi_w) = \frac{\frac{1}{m}\sum_{t=1}^{m}  \left(\sum_{j=1}^{|y|} \delta_{t,j} \log \pi_w(y_{t,j}|y_{t,<j}, x_t)\right)}{\frac{1}{n}\sum_{t=1}^{n} \pi_{w'}(y_t|x_t)}.
\end{align}

\FloatBarrier
\begin{figure*}
	\centering
	\includegraphics[width=\textwidth,keepaspectratio]{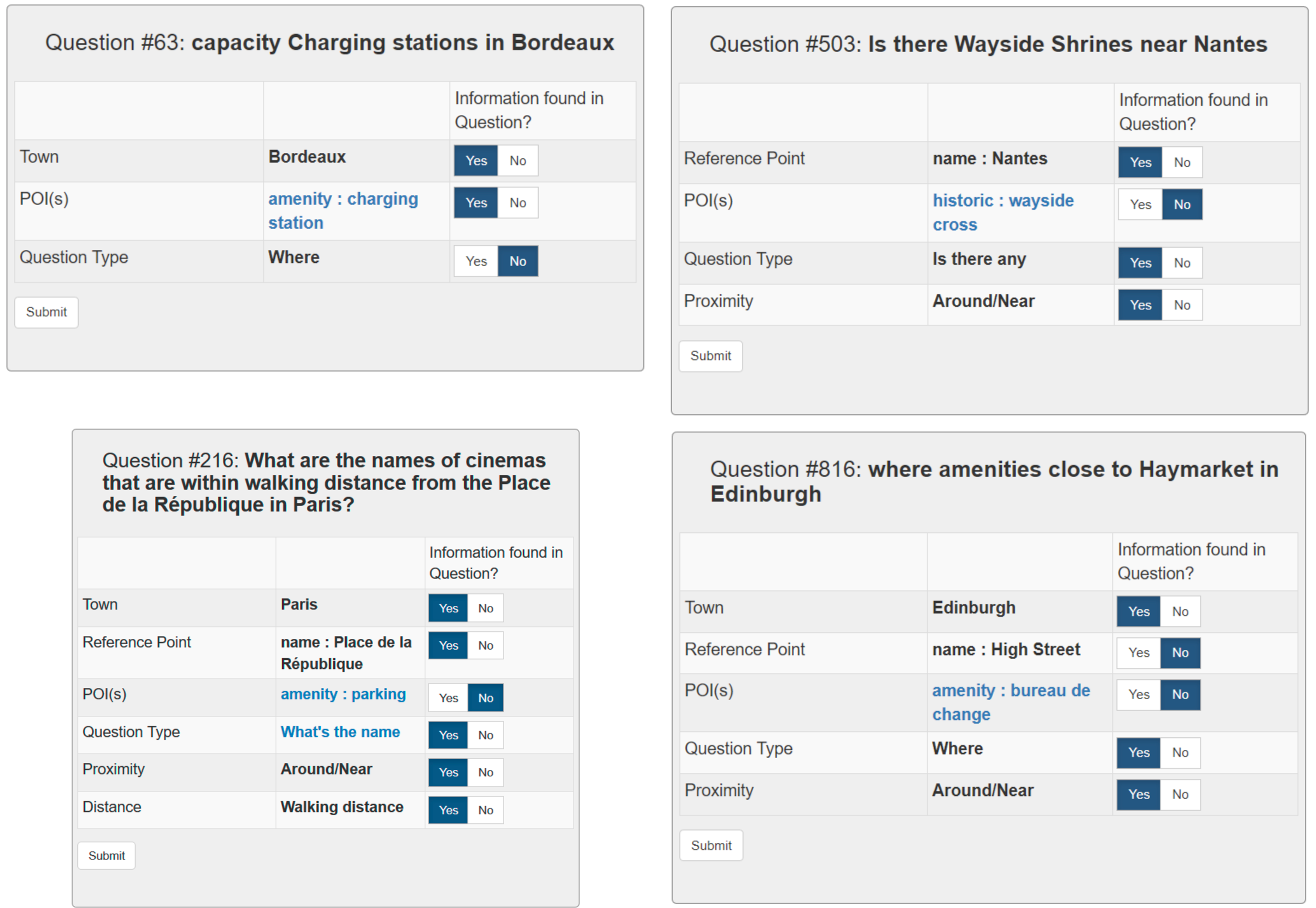}
	\caption{Feedback forms for several questions as filled out by a human user.}
	\label{fig:user_interface2}
\end{figure*}

\section{Feedback Collection}\label{sec:feedback}
OpenStreetMap (OSM) is a geographical database in which volunteers annotate points of interests in the world. A point of interest consists of one or more associated GPS points. Further relevant information may be added at the discretion of the volunteer in the form of tags. Each tag consists of a key and an associated value, for example ``\textit{tourism : hotel}''.
The \textsc{NLmaps v2} corpus was introduced by \citet{LawrenceRiezler:18} as an extension to \textsc{NLmaps} \citep{HaasRiezler:16}. It pairs English questions with machine readable parses, i.e. queries that can be executed against OSM.

\paragraph{Human Feedback Collection.}
The task of creating a natural language interface for OSM demonstrates typical difficulties that make it expensive to collect supervised data. The machine readable language of the queries is based on the \textsc{Overpass} query language which was specifically designed for the OSM database. It is thus not easily possible to find experts that could provide correct queries. It is equally difficult to ask workers at crowdsourcing platforms for the correct answer. For many questions, the answer set is too large to expect a worker to count or list them all in a reasonable amount of time and without errors. For example, for the question ``\textit{How many hotels are there in Paris?}'' there are 951 hotels annotated in the OSM database.
Instead we propose to automatically transform the query into a block of statements that can easily be judged as correct or incorrect by a human. The question and the created block of statements are embedded in a user interface with a form that can be filled out by users. Each statement is accompanied by a set of radio buttons where a user can select either ``\textit{Yes}'' or ``\textit{No}''. For screenshots of four forms filled out by humans see Figure \ref{fig:user_interface2}.

 In total, there are 8 different statement types and each is triggered based on the shape of the query and certain tokens. An overview of the statement types, their triggers and the value a statement will hold, can be found in Table \ref{exp:statement_types}. For example, the token ``\textit{area}'' triggers the statement type ``\textit{Town}''. The statement is then populated with the corresponding information from the query. In the case of ``\textit{area}'', the following OSM value is used, e.g. ``\textit{Paris}''. With this, the meaning of every query can be captured by a block of human-understandable statements.

\begin{table*}
	\begin{center}
		\begin{tabular}{ll}
			\toprule
			Type& Explanation  \\
			\midrule
			Town & OSM tags of ``\textit{area}''\\
			Reference Point &OSM tags ``\textit{center}''\\
			POI(s)&OSM tags of ``\textit{search}'' if ``\textit{center}'' is set,\\
			& else of ``\textit{nwr}''\\
			Question Type &Arguments of ``\textit{qtype}''\\
			Proximity : Around/Near&If ``\textit{around}'' is present\\
			Restriction : Closest&If ``\textit{around}'' and ``\textit{topx(1)}'' are present\\
			Distance&Argument of ``\textit{maxdist}''\\
			Cardinal Direction&``\textit{north}'', ``\textit{east}'', ``\textit{south}'' or ``\textit{west}'' are present\\
			\bottomrule
		\end{tabular}
		\caption{Overview of the possible statements types that are used to transform a parse into a human-understandable block of statements.}
		\label{exp:statement_types}
	\end{center}
\end{table*}

OSM tags and keys are generally understandable. For example, the correct OSM tag for ``\textit{hotels}'' is ``\textit{tourism : hotel}'' and when searching for websites, the correct question type key would be ``\textit{website}''. Nevertheless, for each OSM tag or key, we automatically search for the corresponding Wikipedia page on the OpenStreetMap Wiki\footnote{\url{https://wiki.openstreetmap.org/}} and extract the description for this tag or key. The description is made available to the user in form of a tool-tip that appears when hovering over the tag or key with the mouse. If a user is unsure if a OSM tag or key is correct, they can read this description to help in their decision making. For example, hovering over the tag  ``\textit{amenity : parking}'' will show a small pop-up box with the description: ``\textit{A place for parking cars}''. Once the form is submitted, a script maps each statement back to the corresponding tokens in the original query. These tokens then receive negative or positive feedback based on the feedback the user provided for that statement.

\section{Experiments}\label{sec:exp}

\paragraph{General Settings.}
In our experiments we use the sequence-to-sequence neural network package \textsc{Nematus} \citep{SennrichETAL:17}. Following the method used by \citet{HaasRiezler:16}, we split the queries into individual tokens by taking a pre-order traversal of the original tree-like structure. For example, ``\textit{query(west(area(keyval('name','Paris')), nwr(keyval('railway','station'))),qtype(count))}'' becomes ``\textit{query@2 west@2 area@1 keyval@2 name@0 Paris@s nwr@1 keyval@2 railway@0 station@s qtype@1 count@0}''.

The SGD optimizer used is ADADELTA \citep{Zeiler:12}. The model employs 1,024 hidden units and word embeddings of size 1,000. The maximum sentence length is 200 and gradients are clipped if they exceed a value of 1.0. The stopping point is determined by validation on the development set and selecting the point at which the highest evaluation score is obtained. Validation is run after every 100 updates, and each update is made on the basis of a minibatch of size 80.

The evaluation of all models is based on the answers obtained by executing the most likely query obtained after a beam search with a beam of size 12. We report the F1 score which is the harmonic mean of precision and recall. Recall is defined as the percentage of fully correct answers divided by the set size. Precision is the percentage of correct answers out of the set of answers with non-empty strings.  Statistical significance between models is measured using an approximate randomization test \citep{Noreen:1989}.

\paragraph{Baseline Parser \& Log Creation.} Our experiment design assumes a baseline neural semantic parser that is trained in fully supervised fashion using a cross-entropy objective, and is to be improved by bandit feedback obtained for system outputs from the baseline system for given questions. For this purpose, we select 2,000 question-query pairs randomly from the full extended \textsc{NLmaps v2} corpus. We will call this dataset $\D_{sup}$. Using this dataset, a baseline semantic parser is trained in supervised fashion. It obtains an F1 score of 57.45\% and serves as the logging policy $\pi_0$.

Furthermore we randomly split off 1,843 and 2,000 pairs for a development and test set, respectively. This leaves a set of 22,765 question-query pairs. The questions can be used as input and bandit feedback can be collected for the most likely output of the semantic parser. We refer to this dataset as $\D_{log}$.

\paragraph{Collection of Human Bandit Feedback.}
To collect human feedback, we take the first 1,000 questions from $\D_{log}$ and use $\pi_0$ to parse these questions to obtain one output query for each. 5 question-query pairs are discarded because the suggested query is invalid. For the remaining question-query pairs, the queries are each transformed into a block of human-understandable statements and embedded into the user interface described in Section \ref{sec:feedback}.
We recruited 9 users to provide feedback for these question-query pairs. The resulting log is referred to as $\D_{human}$. Every question-query pair is purposely evaluated only once to mimic a realistic real-world scenario where user logs are collected as users use the system. In this scenario, it is also not possible to explicitly obtain several evaluations for the same question-query pair. Some examples of the  received feedback can be found in Figure \ref{fig:user_interface2}.

To verify that the feedback collection is efficient, we measured the time each user took from loading a form to submitting it. To provide feedback for one question-query pair, users took 16.4 seconds on average with a standard deviation of 33.2 seconds. The vast majority (728 instances) are completed in less than 10 seconds.

\paragraph{Learning from Human Bandit Feedback.}
An analysis of $\D_{human}$ shows that for 531 queries all corresponding statements were marked as correct. We consider a simple baseline that treats completely correct logged data as a supervised data set with which training continues using the cross-entropy objective. We call this baseline bandit-to-supervised conversion (B2S).
Furthermore, we present experimental results using the log $\D_{human}$ for stochastic (minibatch) gradient descent optimization of the counterfactual objectives introduced in equations \ref{eq:emp-risk}, \ref{eq:prob_mini}, \ref{eq:token} and \ref{eq:token_prob}. For the token-level feedback, we map the evaluated statements back to the corresponding tokens in the original query and assign these tokens a feedback of 0 if the corresponding statement was marked as wrong and 1 otherwise. In the case of sequence-level feedback, the query receives a feedback of 1 if all statements are marked correct, 0 otherwise. For the OSL objectives, a separate experiment (see \citet{LawrenceRiezler:18}) showed that updating the reweighting constant after every validation step promises the best trade-off between performance and speed.

Results, averaged over 3 runs, are reported in the left half of Table \ref{exp:results}. The B2S model can slightly improve upon the baseline but not significantly. DPM improves further, significantly beating the baseline. Using the multiplicative control variate modified for SGD by OSL updates does not seem to help in this setup.
By moving to token-level rewards, it is possible to learn from partially correct queries. These partially correct queries provide valuable information that is not present in the subset of correct answers employed by the previous models. Optimizing DPM+T leads to a slight improvement and combined with the multiplicative control variate, DPM+T+OSL yields an improvement of about 1.0 in F1 score upon the baseline. It beats both the baseline and the B2S model by a significant margin.

\begin{table}
		\begin{tabular}{llll|llll}
			\toprule
			\multicolumn{2}{l}{\small\textsc{Human Feedback}}& F1 & $\Delta$ F1 & \multicolumn{2}{l}{\small\textsc{Simulated Feedback}}& F1 & $\Delta$ F1 \\
			\midrule
			\raisebox{.1\height}{\scriptsize 1}&baseline&57.45& & \raisebox{.1\height}{\scriptsize 1}&baseline&57.45&\\
			\raisebox{.1\height}{\scriptsize 2}&B2S&57.79\small$\pm0.18$&+0.34 & \raisebox{.1\height}{\scriptsize 2}&B2S$^{1,3}$&63.22\small$\pm0.27$&+5.77\\
			\raisebox{.1\height}{\scriptsize 3}&DPM$^1$&58.04\small$\pm0.04$&+0.59 & \raisebox{.1\height}{\scriptsize 3}&DPM$^{1}$&61.80\small$\pm0.16$&+4.35\\
			\raisebox{.1\height}{\scriptsize 4}&DPM+OSL&58.01\small$\pm0.23$&+0.56 & \raisebox{.1\height}{\scriptsize 4}&DPM+OSL$^{1,3}$&62.91\small$\pm0.05$&+5.46\\
			\raisebox{.1\height}{\scriptsize 5}&DPM+T$^1$&58.11\small$\pm0.24$&+0.66 & \raisebox{.1\height}{\scriptsize 5}&DPM+T$^{1,2,3,4}$&63.85\small$\pm0.2$&+6.40\\
			\raisebox{.1\height}{\scriptsize 6}&DPM+T+OSL$^{1,2}$&58.44\small$\pm0.09$&+0.99 & \raisebox{.1\height}{\scriptsize 6}&DPM+T+OSL$^{1,2,3,4}$&64.41\small$\pm0.05$&+6.96\\
			\bottomrule
		\end{tabular}
\bigskip
\caption{Answer F1 scores on the test set for the various setups using human feedback (left) or simulated feedback (right), averaged over 3 runs. Statistical significance of system differences at $p<0.05$ are indicated by experiment number in superscript.}
\label{exp:results}
\end{table}

\paragraph{Learning from Large-Scale Simulated Feedback.}
We want to investigate whether the results scale if a larger log is used. Thus, we use $\pi_0$ to parse all 22,765 questions from $\D_{log}$ and obtain for each an output query. For sequence level rewards, we assign feedback of $1$ for a query if it is identical to the true target query, 0 otherwise. We also simulate token-level rewards by iterating over the indices of the output and assigning a feedback of $1$ if the same token appears at the current index for the true target query, 0 otherwise.

An analysis of $\D_{log}$ shows that 46.27\% of the queries have a sequence level reward of $1$ and are thus completely correct. This subset is used to train a bandit-to-supervised (B2S) model using the cross-entropy objective.
Experimental results for the various optimization setups, averaged over 3 runs, are reported in the right half of Table \ref{exp:results}.
We see that the B2S model outperforms the baseline model by a large margin, yielding an increase in F1 score by 6.24 points. Optimizing the DPM objective also yields a significant increase over the baseline, but its performance falls short of the stronger B2S baseline. Optimizing the DPM+OSL objective leads to a substantial improvement in F1 score over optimizing DPM but still falls slightly short of the strong B2S baseline.
Token-level rewards are again crucial to beat the B2S baseline significantly. DPM+T is already able to significantly outperform B2S in this setup and DPM+T+OSL can improve upon this further.

\section{Discussion}

In our experiments, we first prepared forms that were then filled out by recruited human users. In the future, we envision to incorporate the feedback form directly into the online natural language interface to OSM\footnote{\url{http://nlmaps.cl.uni-heidelberg.de/}}. At the moment, a semantic parser might parse the question ``\textit{How many hotels are there in Paris?}'' with the tag ``\textit{amenity=restaurant}'' in the parse, rather than the correct ``\textit{tourism=hotel}''. A user would remain ignorant that the parser misunderstood the question and that a wrong answer has been presented. With the feedback form, the user can verify for their own comfort that their question was understood correctly.

Going one step further, the feedback form could be transformed into an interactive experience. Instead of only marking errors, a user could be prompted to correct errors. This could be achieved as follows: If a user marks a statement as incorrect, the semantic parser can automatically traverse the $n$-best list for the highest ranking parse where the incorrect statement does not appear and present the new parse and its answer to the user instead. Alternatively, we can allow people to edit the form and directly provide the correct information. In the case of the question type, any user could easily select the correct one in a drop-down menu of the 4 possible question types\footnote{The possible question types are: counting, requesting location, enquiring about the existence of an object and asking to retrieve the value of a specific OSM key, such as \textit{name} or \textit{website}.} present in the \textsc{NLmaps v2} corpus. In the case of OSM tags, a short list from the $n$-best list could be produced and the user could search this list for a fitting tag. Users with knowledge of OSM tags could even directly enter the correct tag. 

An interactive setup would deliver a better user experience while simultaneously collecting valuable feedback that can be used to improve performance. The ideas presented here could easily be transferred and incorporated into commercial personal assistants. Additionally, this is a step towards ensuring that model decisions are visible and understandable to the user. Having greater transparency in the decision-making of artificially intelligent systems has become an important concern to many users. At the same time, the system can learn from the mistakes it makes, rather than remaining ignorant after a dissatisfied user leaves the platform, because they were not able to easily provide feedback and inform the system of their dissatisfaction. Viable and intuitive user-system interactions ensure a bilateral dialogue from which both sides can profit.

\section{Conclusion}
We introduced a scenario for improving a neural semantic parser from human bandit feedback. In order to avoid complex and costly data annotation for supervised learning, especially in commercial applications where weak feedback can be collected easily in large amounts from users, it is important to ease the interaction of users with a system. We propose to transfer outputs of a semantic parser into blocks of human-understandable statements which enable non-expert users to act as proofreaders who mark errors in the semantic parse. In the vast majority of the cases, a block is filled out in less than 10 seconds, which showcases the efficiency of our method.

Our algorithms are designed for counterfactual learning where we introduce a new objective that can leverage fine-grained feedback to learn from partially correct parses.
Furthermore, we presented a reweighting objective that enables us to perform stochastic gradient optimization in a minibatch settings appropriate for neural networks.
We show that a strong baseline using a bandit-to-supervised conversion can be significantly outperformed by a combination of reweighting and token-level rewards collected from non-expert human users.

\subsubsection*{Acknowledgments}
The research reported in this paper was supported in part by DFG grant RI-2221/4-1.	

\bibliography{ref}
\bibliographystyle{apalike}

\end{document}